\documentclass[conference]{IEEEtran}
\IEEEoverridecommandlockouts
\usepackage{cite}
\usepackage{amsmath,amssymb,amsfonts}
\usepackage{algorithmic}
\usepackage{graphicx}
\usepackage{textcomp}
\usepackage{xcolor}
\usepackage{url}

\def\BibTeX{{\rm B\kern-.05em{\sc i\kern-.025 em b}\kern-.08em
    T\kern-.1667em\lower.7ex\hbox{E}\kern-.125emX}}
\begin{document}

\title{Learning Policies from Human Data for Skat\\
\thanks{}
}

\author{\IEEEauthorblockN{Douglas Rebstock\IEEEauthorrefmark{1},
Christopher Solinas\IEEEauthorrefmark{2}, and Michael Buro\IEEEauthorrefmark{3}}
\IEEEauthorblockA{Department of Computing Science \\
University of Alberta\\
Edmonton, Canada\\
Email: \IEEEauthorrefmark{1}drebstoc@ualberta.ca,
\IEEEauthorrefmark{2}solinas@ualberta.ca,
\IEEEauthorrefmark{3}mburo@ualberta.ca}}

\maketitle

{\let\thefootnote\relax\footnote{\textcopyright 2019 IEEE.  Personal use of this material is permitted.  Permission from IEEE must be obtained for all other uses, in any current or future media, including reprinting/republishing this material for advertising or promotional purposes, creating new collective works, for resale or redistribution to servers or lists, or reuse of any copyrighted component of this work in other works.}

\begin{abstract}
Decision-making in large imperfect information games is difficult.  Thanks to
recent success in Poker, Counterfactual Regret Minimization (CFR) methods have
been at the forefront of research in these games.  However, most of the
success in large games comes with the use of a forward model and powerful
state abstractions.  In trick-taking card games like Bridge or Skat,
large information sets and an inability to advance the simulation without
fully determinizing the state make forward search problematic.  Furthermore,
state abstractions can be especially difficult to construct because the
precise holdings of each player directly impact move values.

In this paper we explore learning model-free policies for Skat from human game
data using deep neural networks (DNN).  We produce a new state-of-the-art
system for bidding and game declaration by introducing methods to a)
directly vary the aggressiveness of the bidder and b) declare games based
on expected value while mitigating issues with rarely observed state-action
pairs. Although cardplay policies learned through imitation are slightly
weaker than the current best search-based method, they run orders of magnitude
faster. We also explore how these policies could be learned directly from
experience in a reinforcement learning setting and discuss the value of
incorporating human data for this task.
\end{abstract}


\section{Introduction} \label{sec:intro}

Decision-making in large imperfect information games can be difficult.
Techniques based on counterfactual regret minimization (CFR) 
\cite{zinkevich2008regret} 
are currently considered state-of-the-art, but a forward model and expert 
abstractions are often required to scale these techniques to larger games.
Some games are simply too large to solve with CFR methods on current hardware.
For instance, in the popular 3-player card game of Skat the size of the information set for the first decision point right after the initial
deal can be as large as $\approx 4.3\cdot10^{9}$. Overall, there are $\approx 4.4\cdot10^{19}$ terminal histories in the
pre-cardplay portion alone and many more when taking cardplay into
account.

The general approach for solving larger games with these methods is to first
abstract the game into a smaller version of itself, solve that, and then map
those strategies back to the original game.  This process implies a
game-specific tradeoff between abstraction size and how well the strategies
computed on the abstraction translate to the real game.  Recent advances in
Poker \cite{moravvcik2017deepstack,brown2018superhuman} highlight the
effectiveness of this approach in some games.  In Skat and most other trick-taking-card games,
however, the values of actions in an information set are highly dependent on
the interactions between cards within a player's hand and the exact cards
which each opponent possesses. This makes it difficult to construct
abstractions that are small enough to use with CFR methods, but expressive
enough to capture the per-card dependencies that are vital to success in the
full game.

Effective search is also difficult in Skat due to the imperfect
information nature of the game. In order to advance the state, the state must
be ``determinized'' from the root information set.
 The current state-of-the-art in Skat uses a combination of open-handed
simulation and a table-based state evaluator learned from human games
\cite{buro2009improving}.  It relies on a forward model to perform the
open-handed simulations and hand-based abstractions to build the state
evaluator used for bidding. Open-handed simulations have been rightfully
criticized across the literature \cite{frank1998search,russell2016artificial}
because they assume that a strategy can take different actions in different
states that are part of the same information set.

In this paper we focus on learning model-free policies for Skat from
human-generated data. Abandoning the use of a forward model for Skat is
complicated, but may be worthwhile not only because it alleviates many of the
aforementioned problems, but also because it allows policies to be trained or
improved directly through experience. In particular, techniques that alleviate
some of the issues with learning from expert data are explored. We present a
method for varying the aggressiveness of the bidder by viewing the output of
the network as a distribution of the actions of the humans and selecting the
action that maps to the percentile of bidder aggression we desire. Imitation
policy performance is further improved by accounting for rarely-seen
state-action pairs without generating new experience. Our contributions lead to 
a new state-of-the-art bidding system for Skat, and a reasonably strong card player
that performs orders of magnitude faster than search based methods. Finally,
we explain how these policies could be learned directly from experience and
discuss the value of incorporating human data into this process.

\section{Background and Related Work} \label{sec:bg}

In this section we provide the reader with the necessary background related to
the game of Skat. We also discuss previous work related to AI systems for Skat
and similar domains.

\subsection{Skat}

Our domain of choice is a 3-player trick-taking card game called Skat.
Originating in Germany in the 1800s, Skat is played competitively in clubs
around the world. The following is a shortened explanation that includes the
necessary information to understand the work presented here.  For more
in-depth explanation about the rules of Skat we refer interested readers to
{\small\url{https://www.pagat.com/schafk/skat.html}}.

Skat is played using a 32-card deck which is built from a standard 52-card by
removing 2,3,4,5,6 in each suit. A hand consists of each of the three
players being dealt 10 cards with the remaining two kept face down in the
so-called skat.

Games start with the bidding phase. The winner of this phase plays as the
soloist against the team formed by the other players during the cardplay phase
of the game. Upon winning the bidding, the soloist decides whether or not to
pickup the skat followed by discarding two cards face down, and then declares
what type of game will be played during cardplay. The game type declaration
determines both the rules of the cardplay phase and also the score for each
player depending on the outcome of the cardplay phase. Players typically play
a sequence of 36 of such hands and keep a tally of the score over all
hands to determine the overall winner.

The game value, which is the number of points the soloist can win, is the
product of a base value (determined by the game type, see
Table~\ref{tab:gametype}) and a multiplier. The multiplier is determined by
the soloist having certain configurations of Jacks and other high-valued
trumps in their hand and possibly many game type modifiers explained in
Table~\ref{tab:modifiers}. An additional multiplier is applied to the game
value for every modifier.

After dealing cards, the player to the right of the dealer
starts bidding by declaring a value that must be less than or equal to the
value of the game they intend to play --- or simply passing. If the soloist
declares a game whose value ends up lower than the highest bid, the game is
lost automatically. Next, the player to the dealer's left decides whether to
accept the bid or pass. If the player accepts the bid, the initial bidder must
proceed by either passing or bidding a higher value than before. This
continues until one of the player's decides to pass. Finally, the dealer
repeats this process by bidding to the player who has not passed. Once two
players have passed, the remaining player has won the bidding phase and
becomes the soloist.
At this point, the soloist decides whether or not to pick up the skat and
replace up to two of the cards in their hand and finally declares a game type.

\begin{table}[t]
  \caption{Game Type Description}
  \label{tab:gametype}
  {
  \small
  \begin{tabular}{cccc}
        & Base  &        & Soloist Win\\ 
   Type & Value & Trumps & Condition\\
    \hline
    Diamonds & 9 & Jacks and Diamonds & $\ge$ 61 card points\\
    Hearts & 10 & Jacks and Hearts & $\ge$ 61 card points\\
    Spades & 11 & Jacks and Spades & $\ge$ 61 card points\\
    Clubs & 12 & Jacks and Clubs & $\ge$ 61 card points\\
    Grand & 24 & Jacks & $\ge$ 61 card points\\
    Null & 23 & No trump & losing all tricks\\
  \end{tabular}
  }
\bigskip
  \caption{Game Type Modifiers}
  \label{tab:modifiers}
  {\small
  \begin{tabular}{cl}
    Modifier & Description \\
    \hline
    Schneider & $\ge$90 card points for soloist\\
    Schwarz & soloist wins all tricks\\
    Schneider Announced & soloist loses if card points $<90$\\
    Schwarz Announced & soloist loses if opponents win a trick\\
    Hand & soloist does not pick up the skat\\
    Ouvert & soloist plays with hand exposed\\
  \end{tabular}
  }
  \vspace{-0.5cm}
\end{table}

Cardplay consists of 10 tricks in which the trick leader (either the player
who won the previous trick or the player to the left of the dealer in the
first trick) plays the first card. Play continues clockwise around the table
until each player has played. Passing is not permitted and players must play a
card of the same suit as the leader if they have one --- otherwise any card
can be played. The winner of the trick is the player who played the highest
card in the led suit or the highest trump card.

In suit and grand games, both parties collect tricks which contain point cards
(Jack:2, Queen:3,King:4,Ten:10,Ace:11) and non-point cards (7,8,9). Unless
certain modifiers apply, the soloist must get 61 points or more out of the
possible 120 card points in the cardplay phase to win the game. In null games
the soloist wins if they lose all tricks.



\subsection{Previous Work}

Previous work on Skat AI has applied separate solutions for decision-making in
the pre-cardplay and cardplay phases. The cardplay phase has received the most
attention --- probably due to its similarity to cardplay in other trick-taking
card games.  

Despite its shortcomings, Perfect Information Monte-Carlo (PIMC)
Search \cite{levy1989million} continues be the state-of-the-art cardplay
method for Skat and other trick-taking card games like Bridge
\cite{ginsberg2001gib} and Hearts \cite{sturtevant2008analysis}. Later,
Imperfect Information Monte-Carlo Search \cite{furtak2013recursive} and
Information Set Monte Carlo Tree Search \cite{cowling2012information} sought
to address some of the issues inherent in PIMC while still relying on the use
of state determinization and a forward model.

The current state-of-the-art for the pre-cardplay phase
\cite{buro2009improving} uses forward search --- evaluating leaf nodes after
the discard phase using the Generalized Linear Evaluation Model (GLEM) framework \cite{Buro1998simple}. The
evaluation function is based on a generalized linear model over table-based
features indexed by abstracted state properties. These tables are computed
using human game play data. Evaluations take the player's hand, the game type,
the skat, and the player's choice of discard into account to predict the
player's winning probability. The maximum over all player choices of discard
and game type is taken and then averaged over all possible skats. Finally, the
program bids if the such estimated winning probability is higher than some
constant threshold.

The current strongest Skat AI system, Kermit, utilizes these separate solutions
and plays at human expert strength \cite{buro2009improving}. More recently, the
cardplay strength was greatly improved by using cardplay history to bias the 
sampling in PIMC state determinization \cite{solinas2019improving}. For this paper, 
we used this improved version of Kermit for all experimentation.

\section{Learning Bidding Policies from Human Data} \label{sec:policiesBidding}

In this section we describe the training of the pre-cardplay policies for Skat 
using human data. 
First, we present a simple policy learned through direct imitation of human 
play. 
Next, we study the issue of trying to imitate an action from supervised data 
when intent is not visible.
This is a problem when learning policies for Skat's bidding phase because the
data doesn't show how high the player was willing to bid with their hand. 
Finally, we explore using a value network in conjunction with a policy
for the declaration/pickup phases.

The pre-cardplay phase has 5 decision points: max bid for the Bid/Answer
Phase, max bid for the Continue/Answer Phase, the decision whether to pickup
the skat or declare a hand game, the game declaration and the discard. The
bidding phases feature sequential bids between two players,
but further bids can only be made if the other player has not already
passed. This allows a player to effectively pre-determine what their max bid
will be.  This applies to both the Bid/Answer and Continue/Answer phases.
However, in the Continue/Answer phase remaining players must consider which
bid caused a player to pass in the first bidding phase.  The Declaration and
Discard phases happen simultaneously and could be modelled as a single
decision point, but for simplicity's sake we separate them.

For each decision point, a separate DNN was trained using human data from a popular Skat
server \cite{doskv2018skat}. 
For discard, separate networks were trained for each game type except for Null and
Null Ouvert.
These were combined because of their similarity and the low frequency of Ouvert
games in the dataset.

The features for each network are one-hot encoded. The features and the number of bits
for each are listed in Table~\ref{tab:biddingfeatures}. 
The Bid/Answer network uses the Player Hand, and Player Position features. 
The Continue/Answer network uses the same features as Bid/Answer, plus the Bid/Answer
Pass Bid, which is the pass bid from the Bid/Answer phase. The Hand/Pickup network uses
the same as Continue/Answer, plus the Winning Bid. The Declare network uses the Player Hand +
Skat feature in place of the Player Hand feature, as the skat is part of their hand at
this point. The Discard networks use the same features as the Declare network, with the addition of the Ouvert feature, which indicates whether the game is played open. 

\begin{table}[t]
  \centering
  \caption{Network input features}
  \label{tab:biddingfeatures}
  {
    \small
    \begin{tabular}{c c}
      Features & Width \\
      \hline
      Player Hand & 32 \\
      Player Position  & 3 \\
      Bid/Answer Pass Bid & 67 \\ 
      Winning Bid & 67 \\
      Player Hand + Skat & 32 \\
      Ouvert & 1 \\
    \end{tabular}
  }
  \bigskip
  
  \centering
  \caption{Corresponding actions to Network Outputs}
  \label{tab:biddingoutputs}
  {
    \small        
    \begin{tabular}{c c c}
      Phase & Action & Width \\
      \hline
      Bid/Answer & MaxBid & 67 \\
      Continue/Answer & MaxBid & 67 \\
      Hand/Pickup & Game Type or Pickup & 13 \\	
      Declare & Game Type & 7\\
      Discard & Pair of Cards & 496 \\
    \end{tabular}
  }

  \bigskip

  \centering
  \caption{Pre-cardplay training set sizes and imitation accuracies}
  \label{tab:biddingtraining}
  \vspace{-0.3cm}
  {
    \small
    \begin{tabular}{c r r r}
      & Train Size & Train & Test \\
      Phase & (millions) & Acc.\%  & Acc.\%\\
      \hline
      Bid/Answer    & 23.2   & 83.5 & 83.2  \\
      Continue/Answer & 23.2  & 79.7 & 80.1 \\
      Pickup/Hand   & 23.2   & 97.3 & 97.3 \\
      Declare       & 21.8   & 85.6 & 85.1 \\
      Discard Diamonds & 2.47 & 76.3 & 75.7  \\
      Discard Hearts   & 3.13 & 76.5 & 75.0  \\
      Discard Spades   & 3.89 & 76.5 & 75.5 \\
      Discard Clubs    & 5.07 & 76.6 & 75.8  \\
      Discard Grand    & 6.21 & 72.1 & 70.3  \\
      Discard Null     & 1.46 & 84.5 & 83.2 \\
    \end{tabular}
  }
  \vspace{-0.3cm}
\end{table}

Note that the game type is not included in the feature set of the Discard
networks because they are split into different networks based on that
context. Assuming no skip bids (not normally seen in Skat) these features
represent the raw information needed to reconstruct the game state as
observed from the player. Thus, abstraction and feature engineering in our 
approach is limited.

The outputs for each network correspond to any of the possible actions in the
game at that phase. The legality of the actions depend on the state.
Table~\ref{tab:biddingoutputs} lists the actions that correspond to the
outputs of each network, accompanied by the number of possible actions.

The networks all have identical structure, except for the input and output
layers. Each network has 5 fully connected hidden layers
with RELU \cite{nair2010rectified} activation gates.  The network structure
can be seen in Figure~\ref{fig:network}. The largest of the pre-cardplay networks consists of just over 2.3 million weights.  Tensorflow
\cite{abadi2016tensorflow} was used for the entire training pipeline.
Networks are trained using the ADAM optimizer \cite{kingma2014adam} to
optimize cross-entropy loss with a constant learning rate set to $10^{-4}$.
The middle 3 hidden layers incorporate Dropout \cite{srivastava2014dropout},
with keep probabilities set to 0.6. Dropout is only used for learning on the training set. Each network was trained with early
stopping \cite{prechelt1998automatic} for at most 20 epochs. The size of the training sets, and accuracies
after the final epoch are listed for each network in
Table~\ref{tab:biddingtraining}.  These accuracies appear to be quite
reasonable, given the number of options available at each decision point.  The
test dataset sizes were set to 10,000. Training was done using a single GPU 
(Nvidia GTX 1080 Ti) and took around 8 hours for the training of all pre-cardplay
finalized networks.

\begin{figure*}
  \centering
  \includegraphics[width=0.8\textwidth]{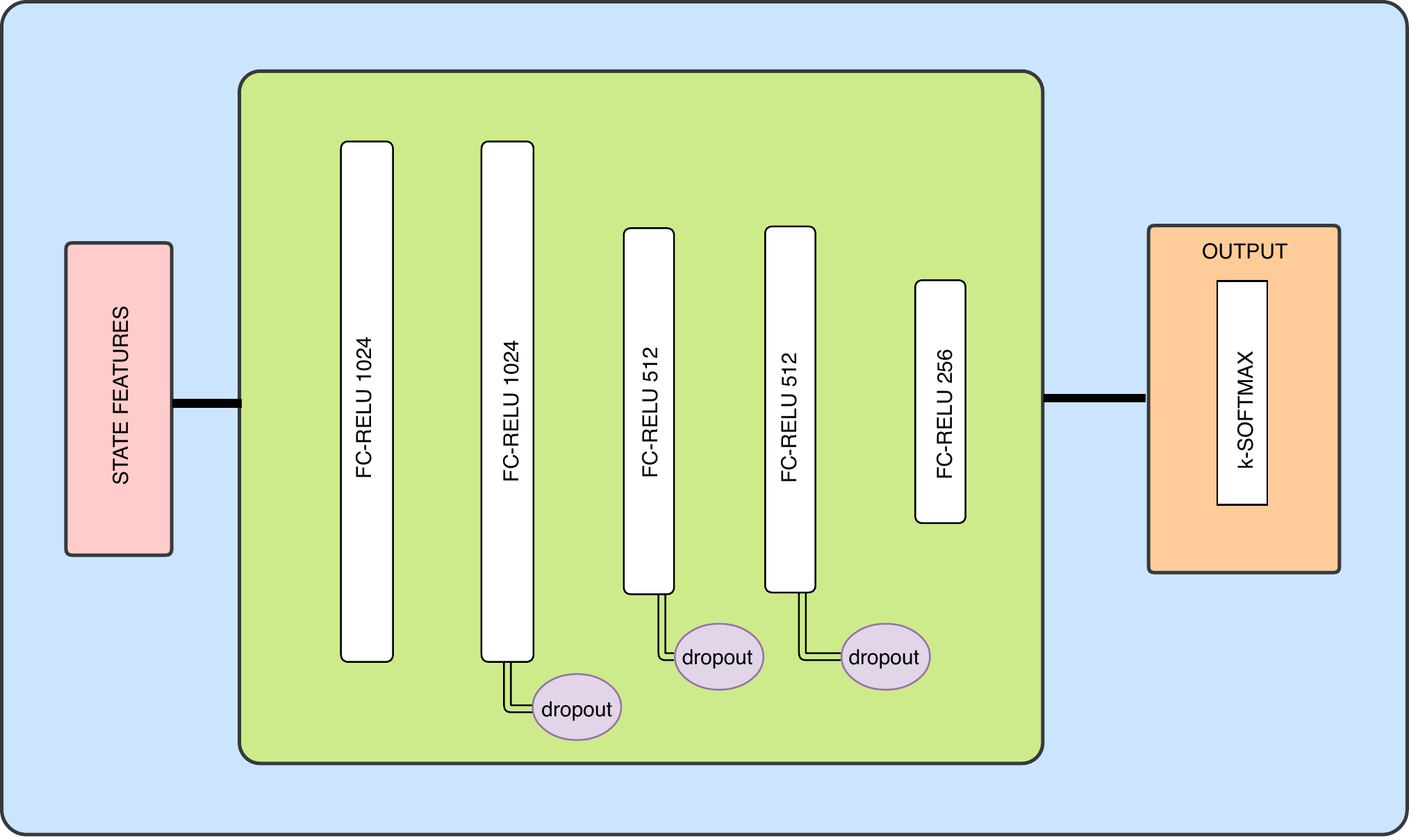}
  \caption{Network architecture used across all game types for both soloist and
  defenders.}
  \label{fig:network}
\end{figure*}

The goal of these networks is to imitate the human players. One issue is that
while the exact actions during the bidding phase are captured, the intent of
how high the player would have bid is not. The intent is based largely on the
strength of the hand, but how high the player bids is dependent on the
what point the other player passed. For example, if a player decided their
maximum bid was 48 but both other players passed at 18, the maximum bid
reached is 18. For this reason, the max bid was trained on the pass bids of
the players. We know what the max bid for these players is because they either
passed at that bid (if they are the player to bid) or at the next bid (if they
are the player to answer).

The output of this network corresponds to the probability the player should
pass at a certain level. The argmax would provide the most likely maxbid, but it would not utilize the
sequential structure of the bids. What we propose is to take the bid that
corresponds to a given percentile, termed $A$. In this way, the distribution
of players aggressiveness in the human population can be utilized directly to
alter the aggressiveness of the player. Let $B$ be the ordered set of possible
bids, with $b_i$ being the $i^{th}$ bid, with $b_0$ corresponding to passing
without bidding. The maxbid is determined using
\begin{equation}
 \text{maxbid}(s,A) = \text{min}(b_i)  \text{~s.t.~}  \Sigma_{j = 0}^{i} p(b_j;\theta | I) \geq A \\
\end{equation}
where $p(b_j)$ is the output of the trained network, which is interpreted as the probability of selecting $b_j$ as the maxbid for the given Information Set $I$ and parameters $\theta$ in the trained network.
Given the returned maxbid is $b_i$ and the current highest bid in the game is $b_{curr}$, the policy for the bidding player is 
\begin{equation}
  \pi_{bid}(b_i,b_{curr}) = \left\{
     \begin{array}{@{}ll}
       b_{current+1}  & \text{if } b_i > b_{current}  \\
       \text{pass} & \text{otherwise} \\
     \end{array}
   \right.
\end{equation} 
while the policy for the answer player is
\begin{equation}
  \pi_{ans}(b_i,b_{curr}) = \left\{
     \begin{array}{@{}ll}
       \text{yes}  & \text{if } b_i \geq b_{current} \\
       \text{pass} & \text{otherwise} \\
     \end{array}
   \right.
\end{equation}                  

Another limiting factor in the strength of direct imitation is that the policy is
trained to best copy humans, regardless of the strength of the move. While the
rational player would always play on expectation, it appears
there is a tendency for risk aversion in the human data set.
For example, the average human seems to play far fewer Grands than Kermit.
Since Grands are high risk / high reward, they are a 
good indication of how aggressive a player is.  
To improve the pre-cardplay policy in the Hand/Pickup and Declare phases,
we can instead select actions based on learned values for each possible game
type.
Formally, this policy is
\begin{equation}
    \label{eqn:mv}
	\pi_{MV}(I,a;\theta) = argmax(v(I,a;\theta))
\end{equation}
where $v$ is the output of the trained network for the given Information Set $I$, action $a$, pair, and parameters $\theta$, and is interpreted as the predicted value of the game.

We trained two additional networks, one for the Hand/Pickup phase and one for
the Declare phase. These networks were identical to the previous ones, except
linear activation units are used for the outputs. The network was trained to
approximate the value of the actions. The value labels are simply the endgame
value of the game to the soloist.  The loss was the mean squared error of
prediction on the actual action taken.  For the Hand/Pickup network, the train
and test loss were 607 and 623 respectively.  For the Declare network, the
values were 855 and 898. These values seem quite large, but with the high
variance and large scores in Skat, they are in the
reasonable range.

The $\pi_{MV}$ seen in Equation~\ref{eqn:mv} is problematic. The reason for
this is that a lot of actions, while legal, are never seen in the training data
within a given context. This leads to action values that are meaningless,
which can be higher than the other meaningful values.  For example, Null
Ouvert is rarely played, has high game value and is most often won. Thus the
network will predict a high value for the Null Ouvert action in unfamiliar
situations which in turn are not appropriate situations to play Null
Ouvert. This results in an overly optimistic player in the face of uncertainty,
which can be catastrophic. This is demonstrated in the results section.

To remedy this issue, we decided to use the supervised policy network in
tandem with the value network. The probability of an action from the policy
network is indicative of how often the move is expected to be played given the
observation. The higher this probability is, the more likely we have seen a
sufficient number of ``relevant'' situations in which the action was
taken. With a large enough dataset, we assume that probabilities above a
threshold indicate that we have enough representative data to be confident in
the predicted action value. We chose 0.1 as the threshold. There is no
theoretical reason for this exact threshold, other than it is low enough that
it guarantees that there will always be a value we are confident in. Furthermore, the
probability used is normalized after excluding all illegal actions.

The policy for the Hand/Pickup and Declare phases using the method described above is
\begin{equation}
\pi_{MLV}(I,a;\theta) = argmax(v_{L}(I,a;\theta)) \\
\end{equation}
where
\vspace{-0.3cm}
\begin{equation}
  v_L(I,a;\theta) = \left\{
     \begin{array}{@{}ll}
	v(I,a;\theta)  & \text{if }p_{legal}(a;\theta|I) \geq \lambda \\
       -\infty & \text{otherwise} \\
     \end{array}
   \right.
\end{equation}
in which $p_{legal}$ is the probability normalized over all legal actions and
$\lambda$ is a constant set to 0.1 in our case.

\section{Bidding Experiments} \label{sec:resultsBidding}

Since Kermit is the current strongest Skat AI system\cite{buro2009improving}
\cite{solinas2019improving}, it is used as the baseline for the rest of this 
paper is Kermit \cite{buro2009improving}. Since the network-based
player learned off of human data, it is assumed that defeating Kermit is
indicative of the overall strength of the method, and not based on exploiting
it.

Because Skat is a 3-player game, each match in the tournament is broken into
six games. In a match, all player configurations
are considered, with the exception of all three being the same bot,
resulting in six games (see Table~\ref{tab:matches}).  In each game, once the
pre-cardplay phase is finished, the rest of the game is played out using
Kermit cardplay, an expert level player based on PIMC search which samples 160
worlds --- a typical setting. The results for each bot is the resultant
average over all the games played. Each tournament was ran for 5,000
matches. All tournaments featured the same identical deals in order to
decrease variance.

Different variations of the pre-cardplay policies were tested against the
Kermit baseline. Unless otherwise stated, the policies use the aggressiveness
transformation discussed in the previous section, with the A value following
the policies prefix. The variations are:

\begin{itemize}
\item Direct Imitation Max {\bf(DI.M)}: selects the most probable action from the
  imitation networks
\item Direct Imitation Sample {\bf(DI.S)}: like DI.M, but samples instead of taking the argmax
\item Aggressive Bidding {\bf(AB)}: like DI.M, but uses the aggressiveness
  transformation in bidding
\item Maximum Value {\bf(MV)}: like AB, but selects the maximum value action
  in the Hand/Pickup and Declare phases
\item Maximum Likely Value {\bf(MLV)}: like AB but uses the maximum likely value
  policy, $\pi_{MLV}$, in the Hand/Pickup and Declare phases
\end{itemize}

\begin{table}[t]
  \centering
  \caption{Player configurations in a single match consisting of six hands (K=Kermit, NW=Network Player)}
  \label{tab:matches}
  \small
  \begin{tabular}{c c c c}
    Game Number & Seat1 & Seat2 & Seat3 \\
    \hline
	1 & K & K & NW \\
	2 & K & NW & K \\
	3 & K & NW & NW \\
	4 & NW & NW & K \\
	5 & NW & K & NW \\
	6 & NW & K & K \\
  \end{tabular}
  \vspace{-0.3cm}
\end{table}

\noindent
While the intuition behind the aggressiveness transformation is
rooted in increasing the aggressiveness of the bidder, the choice for A is not
obvious. MLV and AB were investigated with A values of 0.85, 0.89, 0.925. Through limited trial and error, these values were chosen to
approximately result in the player being slightly less aggressive, similarly
aggressive, and more aggressive than Kermit's bidding,
as measured by share of soloist games played in the tournament setting. MV was
only tested with A of 0.89 since it was clear that the issues of overoptimism
were catastrophic.

An overview of the game type selection breakdown is presented in
Table~\ref{tab:biddingbreakdown}, while an overview on the performance is
presented in Table~\ref{tab:biddingresults}. To measure game playing
performance we use the Fabian-Seeger tournament point (TP) scoring system
which awards the soloist (50 + game value) points if they win. In the case of a
loss, the soloist loses (50 + 2$\cdot$ game value) points and the defenders
are awarded 40 points. All tournament points per game {\bf(TP/G)} difference
values reported were found to be significant, unless otherwise stated. These tests were done using pairwise TTests, with a significance level set to p=0.05.

\begin{table}[t]
  \centering
  \caption{Game type breakdown by percentage for each player, over their 5,000
    match tournament. First player soloist games are broken down into
    types. Defense games (Def) and games that were skipped due to all players
    passing (Pass) are also included. The K vs X entries list breakdowns of
    Kermit playing against player(s) X with identical bidding behavior. }
  \label{tab:biddingbreakdown}
  \setlength{\tabcolsep}{5pt}
  {
    \small
    \begin{tabular}{lrrrrrr}
      Match & Grand & Suit & Null & NO & Def & Pass\\
      \hline
DI.S & 6.8 & 17.1 & 1.2 & 0.8 & 68.9 & 5.2\\
DI.M & 6.6 & 16.0 & 0.9 & 0.9 & 69.3 & 6.3\\
MV.89 & 8.2 & 25.4 & 0.0 & 0.3 & 64.8 & 1.4\\
MLV.85 & 10.6 & 19.0 & 1.6 & 1.3 & 65.6 & 1.9\\
MLV.89 & 11.0 & 19.8 & 1.7 & 1.4 & 64.8 & 1.4\\
MLV.925 & 11.6 & 20.5 & 1.8 & 1.5 & 63.6 & 1.1\\
AB.85 & 8.7 & 21.5 & 1.1 & 1.2 & 65.6 & 1.9\\
AB.89 & 9.2 & 22.3 & 1.1 & 1.2 & 64.8 & 1.4\\
AB.925 & 9.7 & 23.1 & 1.2 & 1.3 & 63.6 & 1.1\\
\hline
K vs DI.S & 10.7 & 21.0 & 3.8 & 2.0 & 57.1 & 5.5\\
K vs DI.M & 10.7 & 21.6 & 3.8 & 2.0 & 55.5 & 6.4\\
K vs *.85 & 11.0 & 17.5 & 2.5 & 1.8 & 64.9 & 2.4\\
K vs *.89 & 10.9 & 16.8 & 2.2 & 1.8 & 66.4 & 1.9\\
K vs *.925 & 11.0 & 15.8 & 2.0 & 1.7 & 68.0 & 1.5\\
    \end{tabular}
   }

  \bigskip
  
  \centering
  \caption{Tournament results over 5,000 matches between learned cardplay
    polices and the baseline player (Kermit). All players use Kermit's
    cardplay. Rows are sorted by score difference (TP/G=tournament points per
    game, S=soloist percentage) Starred diff. values were not found to be significant.}
  \label{tab:biddingresults}
  {
    \small
    \setlength{\tabcolsep}{4pt}
    \begin{tabular}{l|rrr|rrr}
      Player (P) & TP/G(P) & TP/G(K) & diff. & S(P) & S(K) & diff. \\
      \hline
MV.89 & -41.29 & 34.29 & -75.59 & 33.8 & 31.7 & 2.11\\
DI.S & 19.17 & 23.31 & -4.14 & 25.9 & 37.2 & -11.30\\
DI.M & 20.61 & 22.70 & -2.10 & 24.3 & 38.1 & -13.77\\
AB.925 & 22.41 & 22.58 & -0.18* & 35.3 & 30.5 & 4.85\\
AB.89 & 22.87 & 22.50 & 0.37* & 33.8 & 31.7 & 2.11\\
AB.85 & 22.97 & 22.56 & 0.41* & 32.5 & 32.8 & -0.27\\
MLV.925 & 23.04 & 22.29 & 0.75 & 35.3 & 30.5 & 4.85\\
MLV.89 & 23.28 & 22.31 & 0.97 & 33.8 & 31.7 & 2.11\\
MLV.85 & 23.59 & 22.42 & \textbf{1.17} & 32.5 & 32.8 & -0.27\\
    \end{tabular}
  }
  \vspace{-0.3cm}
\end{table}

Clearly, naively selecting the max value (MV) in the Hand/Pickup and Declare
phases cause the bot to perform very poorly as demonstrated by it performing
-75.59 TP/G worse than the Kermit baseline. It plays 96$\%$ of its suit games as
hand, which is extremely high to the point of
absurdity. The reason for this overoptimism was discussed already in the
previous section, and these results bear this out.

Direct Imitation Argmax (DI.M) performed much better, but still performed 
slightly worse than the baseline by 2.10 TP/G. Direct Imitation Sample (DI.S) 
performed 4.14 TP/G worse than baseline, slightly worse than DI.M.  The issue 
with being overly conservative is borne
out for both these players with the player being soloist approximately half as 
often as Kermit.

The direct imitation with the aggressiveness transformation (AB) performed better 
than Kermit for the lower values, but slightly worse for AB.925. None of these values were statistically significant. The best value 
for A was 0.85 (AB.85) which leads to +0.41
TP/G against Kermit.  The advantage decrease with increasing A values. At the
0.85 A value, the player is soloist a fewer of 1.81 times per 100 games
played, indicating it is a less aggressive bidder than Kermit.

The players selecting the max value declarations within a confidence threshold
(MLV) performed the best overall, outperforming the AB players at each A value
level. The best overall player against Kermit is the MLV.85 player. It
outperforms Kermit by 1.17 TP/G, 0.76 TP/G more than the best AB player.

The actual breakdown of games is quite interesting, as it shows that the AB
and MLV players are drastically different in their declarations. Across the
board, AB is more conservative as it plays more Suit games and less Grand
games (worth more and typically more risky) than the corresponding MLV
player. Kermit falls somewhere in between. One other trend is that as the A
values increase, the share of soloist games increases, but the majority of the
extra games are Suit. There is a diminishing returns in the number of high
value Grand games.

These results indicate that the MLV method that utilizes the networks train on human data provides the new state-of-the-art for Skat bots in pre-cardplay.

\section{Learning Cardplay Policies} \label{sec:policiesCardplay}

\begin{table}[t]
  \centering
  \caption{Network input features}
  \label{tab:cardplayfeatures}
   {
     \small          
     \begin{tabular}{c c}
       Common Features & Width \\
       \hline
       Player Hand & 32 \\
       Hand Value & 1 \\
       Played Cards (Player, Opponent 1\&2) & 32*3 \\
       Lead Cards (Opponent 1\&2) & 32*2 \\
       Sloughed Cards (Opponent 1\&2) & 32*2 \\
       Void Suits (Opponent 1\&2) & 5*2 \\
       Current Trick & 32 \\
       Trick Value & 1 \\
       Max Bid Type (Opponent 1\&2) & 6*2 \\
       Soloist Points & 1 \\
       Defender Points & 1 \\
       Hand Game & 1 \\
       Ouvert Game & 1 \\
       Schneider Announced & 1 \\
       Schwarz Announced  & 1 \\
       \\
       Soloist Only Features & Width \\
       \hline
       Skat & 32 \\
       Needs Schneider & 1 \\
       \\
       Defender Only Features & Width \\
       \hline
       Winning Current Trick & 1\\
       Declarer Position & 2 \\
       Declarer Ouvert & 32 \\
       \\
       Suit/Grand Features & Width \\
       \hline
       Trump Remaining & 32 \\
       \\
       Suit Only Features & Width \\
       \hline
       Suit Declaration & 4 \\
     \end{tabular}
   }

  \bigskip
  
  \centering
  \caption{Cardplay train/test set sizes}
  \label{tab:cardplaytrain}
  {
    \small          
    \begin{tabular}{c r c c}
      & Train Size &  Train & Test\\
      Phase & (millions) & Acc.\% & Acc.\% \\
      \hline
      Grand Soloist  &   53.7 & 80.3 & 79.7\\
      Grand Defender &  105.4 & 83.4 & 83.1\\
      Suit Soloist   &  145.8 & 77.5 & 77.2\\
      Suit Defender  &  289.1 & 82.3 & 82.2\\
      Null Soloist   &    5.4 & 87.3 & 86.3\\
      Null Defender  &   10.7 & 72.9 & 71.6\\
    \end{tabular}
  }
  \vspace{-0.6cm}
  
\end{table}

With improved pre-cardplay policies, the next step was to create a cardplay
policy based off the human data. To do this, a collection of networks were trained to
imitate human  play using the same network architecture used for the pre-cardplay
imitation networks. Six networks were trained in all; defender and soloist versions of Grand, Suit, and Null.

To capture the intricacies of the cardplay phase, we use handcrafted features
--- listed in Table~\ref{tab:cardplayfeatures}.  Player Hand represents all
the cards in the players hand. Hand Value is the sum of the point values of
all cards in a hand (scaled to the maximum possible value). Lead cards
represents all the cards the player led(first card in the
trick). Sloughed cards indicate all the non-Trump cards that the player played
that did not follow the suit.  Void suits indicate the suits which a player
cannot have based on past moves.  Trick Value provides the point value of all
cards in the trick (scaled to the max possible value). Max Bid Type indicates
the suit bid multipliers that match the maximum bid of the opponents. For
example, a maximum bid of 36 matches with both the Diamond multiplier, 9, and
the Clubs multiplier, 12. The special soloist declarations are encoded in Hand
Game, Ouvert Game, Schneider Announced, and Schwartz announced. Skat encodes
the cards placed in the skat by the soloist, and Needs Schneider indicates
whether the extra multiplier is needed to win the game. Both of these are
specific to the soloist networks. Specific to the defenders are the Winning
Current Trick and the Declarer Ouvert features. Winning Current Trick encodes
whether the current highest card in the trick was played by the defense
partner. Declarer Ouvert represents the soloist's hand if the soloist declared
Ouvert. Trump remaining encodes all the trump cards that the soloist does not
possess and have not been played, and is used in the Suit and Grand
networks. Suit Declaration indicates which suit is trump based on the
soloist's declaration, and is only used in the Suit networks. These features
are one-hot encoded, except for Trick Value, Hand Value, Soloist and Defender
points, which are floats scaled between 0 and 1.  The network has 32 outputs
--- each corresponding to a given card.

The resultant data set sizes, and accuracies after the final epoch
are listed in Table~\ref{tab:cardplaytrain}. The test set had a size of 100,000
for all networks. The accuracies are quite high, however, this doesn't mean
much in isolation as actions can be forced, and the number of
reasonable actions is often low in the later tricks. Training was done using a single GPU 
(Nvidia GTX 1080 Ti) and took around 16 hours for the training of the finalized
cardplay networks. The largest of the cardplay networks consists of just under 
2.4 million weights.

\section{Cardplay Results} \label{sec:resultsCardplay}

Bidding policies from the previous section, as well as Kermit bidding, were
used in conjunction with the learned cardplay networks. The cardplay policy
(C) takes the legal argmax of the game specific network's output, and plays
the corresponding card. AB and MLV bidding policies were tested at all three bidding
aggressiveness levels.

Each played against Kermit in the
same tournament setup from the previous section. Again, all TP/G difference
reported were found to be significant. Results are reported in
Table~\ref{tab:cardplayresults}.

The strongest full network player was MLV.85+C, and it outperformed Kermit by 
1.05 TP/G. All the rest of the cardplay network players performed worse than 
Kermit. Like the bidding results, all MLV players performed better than all AB 
players. Kermit's search based cardplay is quite strong, and it appears to be 
stronger than the imitation cardplay, as demonstrated by it outperforming K+C by 2.61 TP/G.
This substantial difference is probably due to the effectiveness of search to play
near perfectly in the later half of the game when a lot is known about the hands of
the players. 
In this match-up, the bidding is the same so the only difference is the 
cardplay. While these results are compelling, it should be noted that further 
investigation into the interplay between the bidding and cardplay policies is 
required to get a better understanding of their strengths. One advantage
the imitation network cardplay has is that its much faster, taking turn at a
constant rate of around 2.5 ms, as compared to Kermit which takes multiple
seconds on the first trick, and an average time of around 650 ms (both using a
single thread on consumer-level CPU).

\begin{table}[t]
  \centering
  \caption{Cardplay tournament results over 5,000 matches between bots using
    pre-cardplay policies from the previous section and the learned cardplay
    policies. All variants were played against the baseline, Kermit
    (TP/G=tournament points per game, S=soloist percentage) All diff values were found to be statiscally significant.} \small
  \setlength{\tabcolsep}{4pt}
  \label{tab:cardplayresults}
  \begin{tabular}{l|ccc|ccc}
    Player (P) & TP/G(P) & TP/G(K) & diff & S(P) & S(K) & diff \\
    \hline
K+C & 21.94 & 24.55 & -2.61 & 31.9 & 31.9 & 0.00\\
AB.89 & 22.42 & 24.48 & -2.05 & 33.8 & 31.7 & 2.11\\
AB.925+C & 22.26 & 24.20 & -1.94 & 35.3 & 30.5 & 4.85\\
AB.85+C & 22.62 & 24.38 & -1.76 & 32.5 & 32.8 & -0.27\\
MLV.925+C & 22.83 & 24.14 & -1.32 & 35.3 & 30.5 & 4.85\\
MLV.89+C & 23.11 & 24.41 & -1.30 & 33.8 & 31.7 & 2.11\\
MLV.85+C & 23.44 & 22.39 & \textbf{1.05} & 32.5 & 32.8 & -0.27\\
  \end{tabular}
  \vspace{-0.3cm}
\end{table}

\section{Conclusion} \label{sec:concl}

In this paper we have demonstrated that pre-cardplay policies for Skat can be
learned from human game data and that it performs much better than Kermit's
pre-cardplay --- the prior state-of-the-art. Naively imitating all aspects of
the pre-cardplay by taking the argmax over the legal actions (DI-M) resulted
in a bidding policy that performed an average of 1.35 TP/G worse than the
Kermit baseline. The same procedure but with sampling (DI-S) resulted in the
player performing 4.14 TP/G worse than baseline. Using the novel method to
increase the aggressiveness of the bidder led to it performing 0.41 TP/G
better than the baseline, with A set to 0.85 (AB.85). Using this in
conjunction with game declaration based on the predicted values and
probabilities of actions (MLV.85), resulted in the best overall pre-cardplay
policy, beating the baseline by 1.17 TP/G. Also, the time for pre-cardplay
decisions are much faster, as it does not rely on search.

The direct imitation cardplay policy decreases the strength of the overall
player, performing 2.79 TP/G worse than the Kermit player when utilizing the
Kermit bidder. The best overall full network based player was MLV.925+C, which
outperformed Kermit by 1.05 TP/G. This full network player is order of
magnitudes faster than the search based player, and in this tournament setup,
performs better. One drawback is that while more computation can be done to
improve the search (improving the number of worlds sampled for example), the
same cannot be done for the network player.

\subsection{Future Work}

Now that we have established some degree of success training model-free
policies from human data in Skat, the next logical step is to improve these
policies directly through experience similar to the process shown in the
original AlphaGo \cite{silver2016mastering}.
In \cite{srinivasan2018actor}, regret minimization techniques often used to
solve imperfect information games are shown to be related to model-free multi-agent
reinforcement learning. The resulting actor-critic style agent showed fast
convergence to approximate Nash equilibria during self-play in small variants
of Poker.  Applying the same approach may be difficult because of Skat's size
and the fact that it is not zero-sum, but starting with learning a best
response to the full imitation player discussed in this work should be
feasible and may yield a new state-of-the-art player for all phases of the
game.  However, a fully-fledged self-play regime for Skat remains as the end
goal of this work.

Learning policies though self-play has shown to yield strategies that are
``qualitatively different to human play'' in other games
\cite{silver2017mastering}.  This could be problematic because Skat involves
cooperation on defense during the cardplay phase.  Human players use
conventions and signals to coordinate and give themselves the best chance of
defeating the soloist.  In order to play well with humans, policies need to
account for these signals from their partner and send their own.  We plan to
explore how continually incorporating labelled data from human games helps
alleviate this problem.

\section*{Acknowledgment}

We acknowledge the support of the Natural Sciences and Engineering Research Council of Canada (NSERC).

Cette recherche a \'et\'e financ\'e par le Conseil de recherches en sciences naturelles et en g\'enie du Canada (CRSNG).

\bibliographystyle{IEEEtran}
\bibliography{refs}

\end{document}